\documentclass[sn-nature,referee]{sn-jnl}

\usepackage{graphicx}%
\usepackage{multirow}%
\usepackage{amsmath,amssymb,amsfonts}%
\usepackage{amsthm}%
\usepackage{mathrsfs}%
\usepackage[title]{appendix}%
\usepackage{xcolor}%
\usepackage{textcomp}%
\usepackage{manyfoot}%
\usepackage{booktabs}%
\usepackage{algorithm}%
\usepackage{algorithmicx}%
\usepackage{algpseudocode}%
\usepackage{listings}%

\unnumbered

\usepackage{ulem}
\usepackage{xcolor}
\usepackage{enumitem}
\usepackage{float}
\usepackage{pdfpages}
\usepackage{booktabs}
\usepackage{dblfloatfix}
\usepackage{tablefootnote}
\usepackage{lineno}
\usepackage[font=normalsize]{caption}  
\usepackage{colortbl}
\usepackage{xspace}
\newcommand{\ie}{i.e.\xspace}

\begin{document}

\title[Article Title]{SubGrapher: Visual Fingerprinting of \linebreak Chemical Structures}

\maketitle

\section{Author list}

Lucas Morin\textsuperscript{1, 2, *}, Gerhard Ingmar Meijer\textsuperscript{1}, Valéry Weber\textsuperscript{1}, Luc Van Gool\textsuperscript{3, 2}, Peter W. J. Staar\textsuperscript{1}

\section{Affiliations}

\textsuperscript{1} IBM Research, Säumerstrasse 4, 8803 Rüschlikon, Switzerland \\
\textsuperscript{2} Department of Information Technology and Electrical Engineering, ETH Zürich, Sternwartstrasse 7, 8092 Zürich, Switzerland \\
\textsuperscript{3} INSAIT, Sofia University St. Kliment Ohridski, Tsarigradsko shose 111R, 1784 Sofia, Bulgaria \\
\noindent{\textsuperscript{*} Corresponding author: \\ Lucas Morin, lum@zurich.ibm.com}

\section{Keywords}

Molecular Fingerprint, Optical Chemical Structure Recognition, Document Understanding, Functional Group Recognition

\section{Abstract}

Automatic extraction of molecules from scientific literature plays a crucial role in accelerating research across fields ranging from drug discovery to materials science. Patent documents, in particular, contain molecular information in visual form, which is often inaccessible through traditional text-based searches. In this work, we introduce SubGrapher, a method for the visual fingerprinting of molecule and Markush structure images. Unlike conventional Optical Chemical Structure Recognition (OCSR) models that attempt to reconstruct full molecular graphs, SubGrapher focuses on extracting fingerprints directly from images. Using learning-based instance segmentation, SubGrapher identifies functional groups and carbon backbones, constructing a substructure-based fingerprint that enables the retrieval of molecules and Markush structures. Our approach is evaluated against state-of-the-art OCSR and fingerprinting methods, demonstrating superior retrieval performance and robustness across diverse molecule and Markush structure depictions. The benchmark datasets, models, and inference code are publicly available.

\subsection{Scientific contribution statement}
SubGrapher introduces a novel approach to convert molecule and Markush structure images directly into fingerprints in a single step, bypassing traditional SMILES or graph reconstruction. It outperforms existing OCSR and fingerprinting methods for substructure detection and structure retrieval across diverse datasets, including Markush structure images. 

\section{Introduction}

Knowledge in chemistry is spread across structured databases and unstructured sources such as scientific journals, patent documents, books, and corporate documents. Integrating molecular structures, properties, and synthesis data into a unified collection would significantly accelerate research in chemistry and materials science \cite{Pyzer-Knapp2025}. Converting unstructured documents to machine-readable formats enables searching within document collections and training foundational models at larger scales \cite{livathinos2025doclingefficientopensourcetoolkit,auer2024doclingtechnicalreport, nassar2025smoldoclingultracompactvisionlanguagemodel}.
A key challenge in this process is extracting molecular structures from patent documents, where they are primarily represented as images rather than machine-readable text. Molecular databases such as PatCID \cite{Morin2024} and SureChEMBL \cite{SureChEMBL15} aim to address this challenge by employing document ingestion pipelines that integrate page segmentation, chemical image classification, and Optical Chemical Structure Recognition (OCSR).

The main component of these pipelines, OCSR, is the process of converting molecular depictions into structured representations, such as SMILES (Simplified Molecular Input Line Entry System) \cite{SMILES} or molecular graphs. Graph-based OCSR methods identify molecular components, including atoms and bonds, and reconstruct their connectivity. These methods rely on rule-based image processing algorithms, as seen in OSRA \cite{OSRA}, Imago \cite{Imago}, and MolVec \cite{MolVec}, or deep-learning models such as MolGrapher \cite{Morin_2023_ICCV} and ChemGrapher \cite{ChemGrapher}. In contrast, sequence-based OCSR methods, such as DECIMER \cite{DECIMER-1} and MolScribe \cite{MolScribe}, utilize vision encoders with autoregressive text decoders to generate SMILES strings.

Despite advances, OCSR methods face challenges with variations in drawing conventions or degraded image quality. Additionally, certain chemical illustrations cannot be represented as SMILES, especially depictions with non-standard graphical elements (see row 3 in \autoref{fig:qualitative_eval}) or Markush structures \cite{SIMMONS2003195}. This limitation is especially significant in patent analysis, where Markush structures are widely used to define broad molecular classes. Moreover, full molecular reconstruction is not always necessary for applications such as database searching or molecular property prediction. In many cases, users are more interested in identifying molecules with specific substructures rather than their complete structures. Furthermore, predictive models often rely on molecular fingerprints derived from SMILES \cite{doi:10.1021/acs.jcim.9b00798}, making SMILES an intermediate representation rather than a final output.

Molecular fingerprints are vectorized representations of molecular structures, commonly used for similarity searches in databases and predictive tasks such as property prediction or drug target identification \cite{Wen2022}. 
Structural key fingerprints, such as MACCS \cite{MACCS} and PubChem fingerprints \cite{PubChem22}, encode as a binary vector the presence or absence of predefined molecular fragments. Among these fragments, functional groups \cite{Ertl2017} form a key subset due to their well-defined chemical properties.
To improve generalization, other approaches avoid relying on predefined fragment libraries. Instead, they generate fragments directly from the molecular graph, which are then hashed and folded into bit vectors. For example, Daylight-style fingerprints \cite{Willett1998} generate fragments by enumerating linear atom–bond paths, while Extended-Connectivity Fingerprints (ECFP) \cite{ECFP} generate fragments by enumerating circular neighborhoods around atoms.
In addition, molecular fingerprints can also be generated using probabilistic techniques like MinHash fingerprints (MHFP) \cite{Probst2018}, or learned from large datasets using deep learning models such as MoLFormer \cite{Ross2022}.

Integrating OCSR and fingerprinting into a single-step process by directly recognizing functional groups from images would efficiently facilitate the extraction of molecular information from literature. Recently, Fan, \textit{et al.}, \cite{fan2025} explored the recognition of a limited subset of functional groups from images as an auxiliary task to enhance OCSR performance. Additionally, other studies \cite{fang2025molparserendtoendvisualrecognition, Zeng2022} have investigated the use of molecule images as a way to learn molecular features for downstream predictive tasks. However, none of these studies have combined visual recognition of functional groups with visual fingerprinting into a unified approach.

\begin{figure*}
    \centering
    \includegraphics[trim={0 4cm 0 4cm}, clip, width=0.95\textwidth]{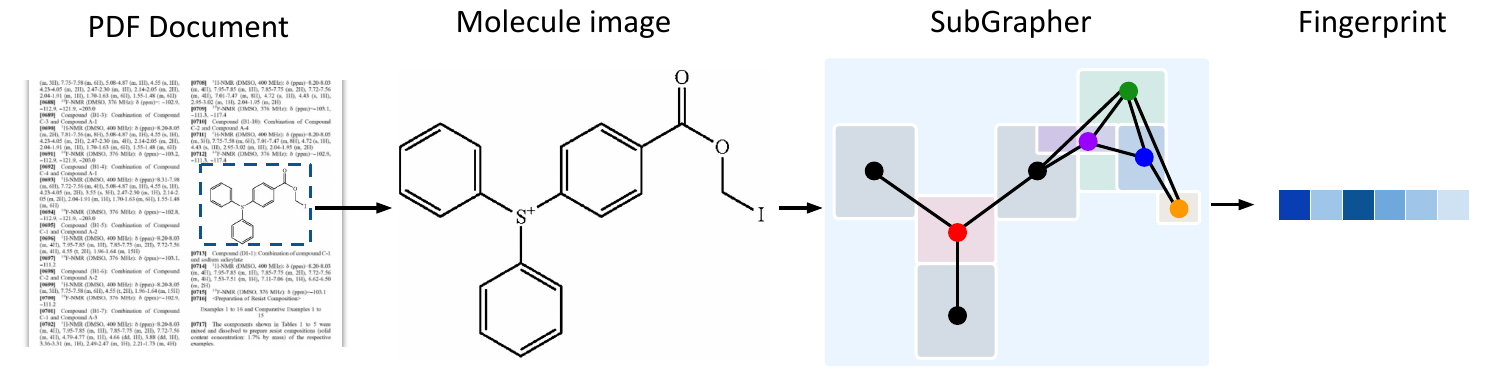}\vspace{5mm}
    \caption{\textbf{SubGrapher} extracts a fingerprint from a molecular or Markush structure image in a document. Our approach identifies functional groups and carbon backbones in images. These substructures are then combined based on their connectivity to create a substructure-graph. Finally, this graph is converted to a fingerprint enabling substructure search, molecule and Markush structure retrieval, or any downstream predictive task.}\vspace{0mm}
    \label{fig:intro}
\end{figure*}

In this work, we introduce SubGrapher for the visual fingerprinting of molecule and Markush structure images, illustrated in \autoref{fig:intro}. First, SubGrapher uses segmentation models to identify functional groups and carbon backbones in images. Second, SubGrapher creates a substructure-graph based on the connectivity of these substructures. Finally, this graph is converted to a count-based continuous fingerprint. On the one hand, our fingerprint allows to search for molecules containing functional groups and carbon backbones of interest. By defining key chemical properties, functional groups are natural targets for molecular searches, while carbon backbones link these functional groups. On the other hand, SubGrapher's fingerprint can be used for molecule and Markush structure retrieval. 

\section{Method}

\begin{figure*}[!t]
    \centering
    \includegraphics[trim={2cm 1cm 2cm 1cm}, clip, width=0.85\textwidth]{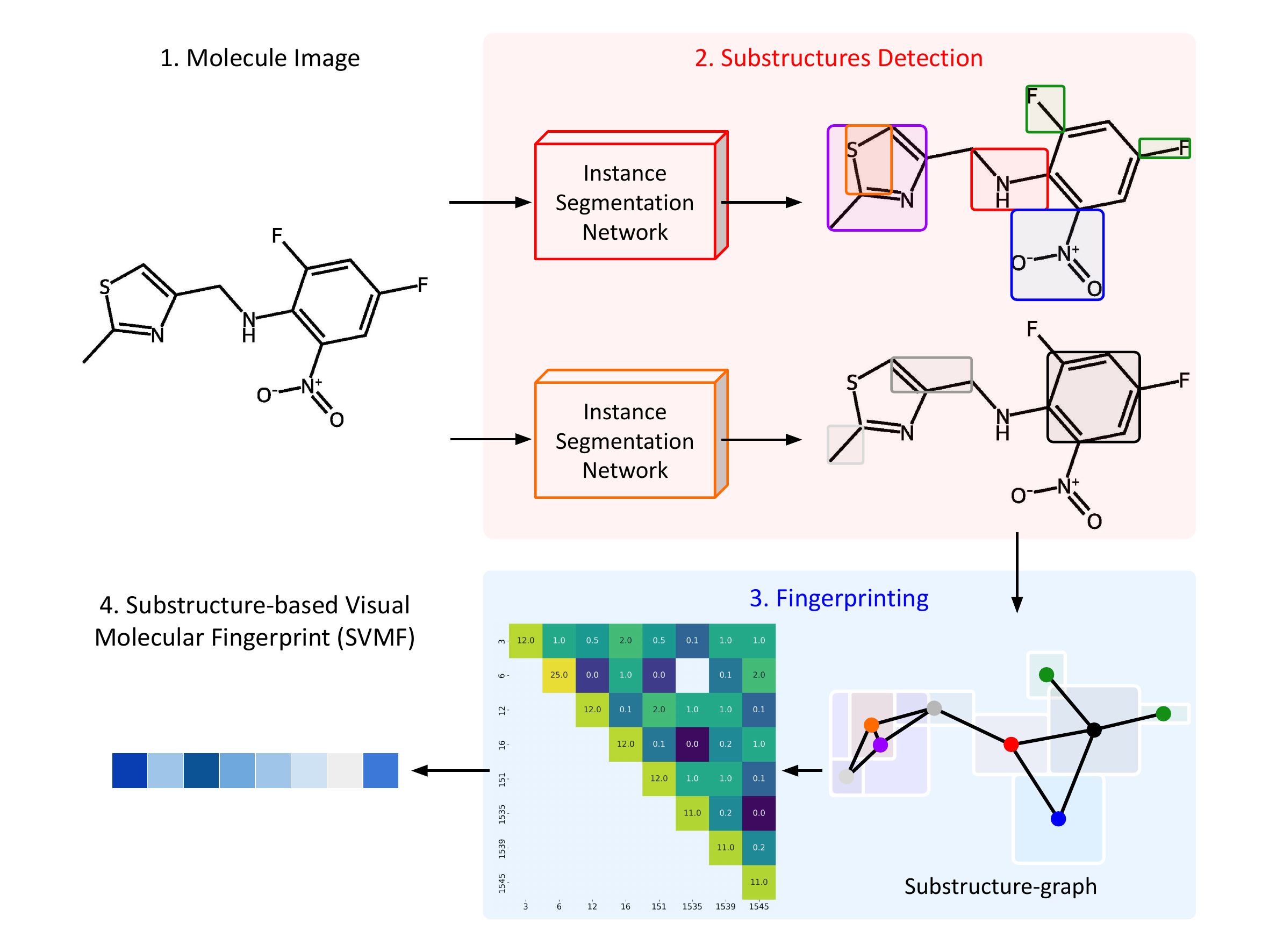}\vspace{5mm}
    \caption{\textbf{SubGrapher architecture.} The SubGrapher model detects functional groups and carbon backbones using instance segmentation networks. The identified substructures are combined into a substructure-graph, which is then converted to a matrix fingerprint. Finally, the matrix is stored as a compressed vector. The resulting fingerprint is count-based, as each coefficient depends on the number of substructure occurrences, and continuous, since the values are real numbers.}\vspace{0mm}
    \label{fig:architecture}
\end{figure*}

We introduce SubGrapher, a model designed for the visual fingerprinting of molecule and Markush structure images. Its architecture is illustrated in \autoref{fig:architecture}.
SubGrapher employs substructure segmentation models to detect and identify molecular substructures, specifically functional groups in the organic chemistry domain and carbon backbones. These detected substructures are then assembled into a substructure-graph and converted into a count-based continuous fingerprint.

\subsection{Substructure Segmentation Model}

To detect molecular substructures from images, SubGrapher employs two segmentation networks, as illustrated in \autoref{fig:architecture}. The first network detects 1534 expert-defined functional groups, defined in the Substructures Coverage section below. The second network identifies 27 distinct carbon backbone patterns, complementing the functional group detection by capturing molecular regions not already assigned to any functional groups. By ensuring that substructures overlap appropriately and span the entire molecule, SubGrapher constructs a more characteristic and informative fingerprint in later stages.

Substructures are detected using mask-based segmentation rather than relying solely on bounding-box detection. This approach provides fine-grained supervision during training, leading to improved accuracy and robustness in molecular substructure recognition. 
In the implementation, the segmentation models are Mask-RCNN \cite{maskrcnn} models.

\subsection{Substructures Coverage}

\begin{figure*}[t]
    \centering
    \includegraphics[trim={0cm 1cm 0cm 1cm}, clip, width=0.9\textwidth]{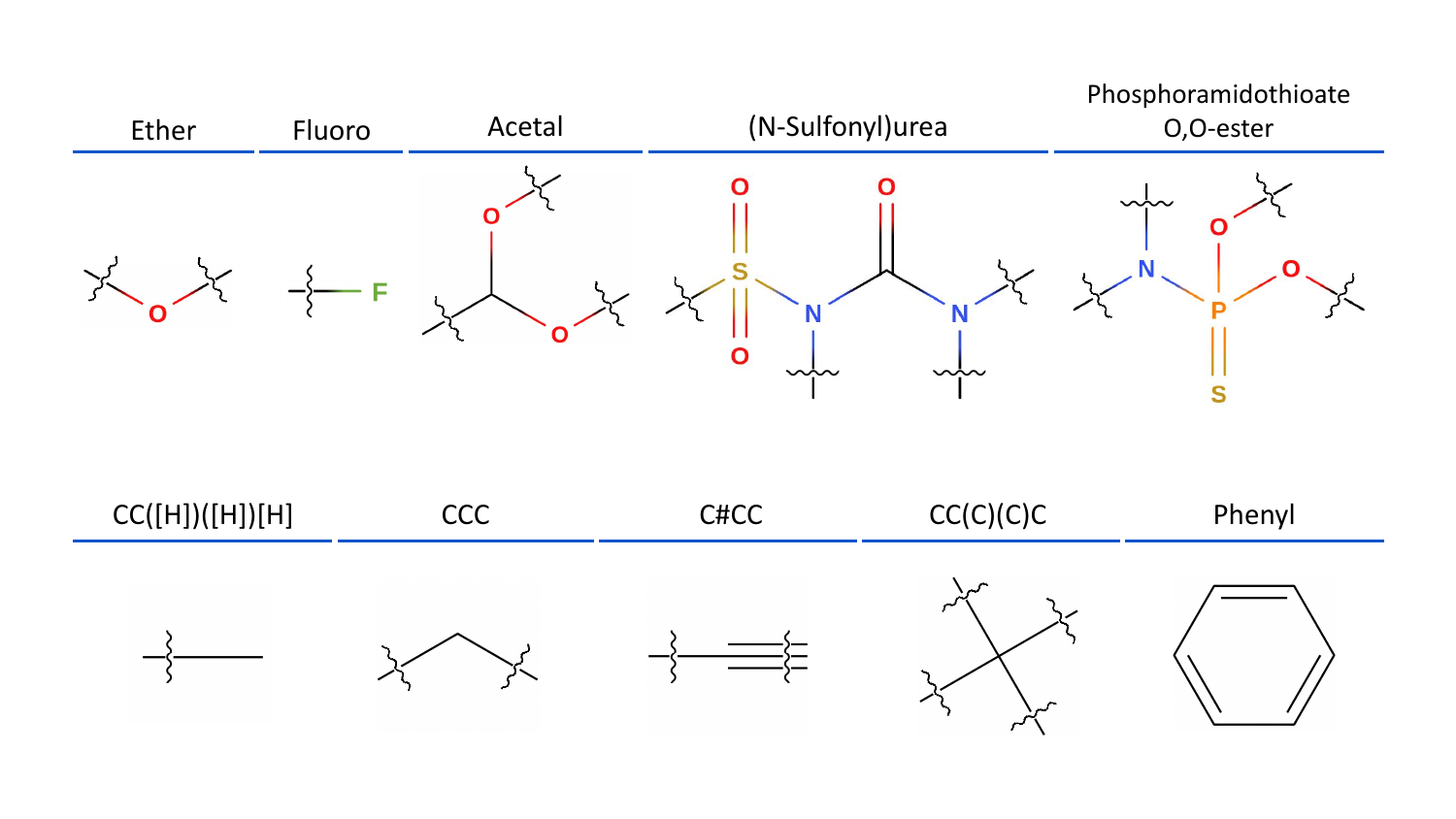}\vspace{1mm}
    \caption{\textbf{Substructure examples.} Examples of functional groups and carbon backbones recognized by SubGrapher.}\vspace{-4mm}
    \label{fig:substructures}
\end{figure*}

Here, we describe the strategy used to select the 1534 functional groups and 27 carbon backbone patterns. Each substructure is manually defined and annotated with a SMILES, SMARTS \cite{smarts}, and a descriptive name.

Our functional groups are defined as substructures containing at least one heteroatom and having for attachment points single bonds connected to carbon atoms. 
These substructures are manually designed starting from chemically logical chains of up to five atoms selected from C, O, S, N, B, or P. 
Each heteroatom in these initial chains can be expanded with chemically relevant subgroups containing atoms C, H, O, S, or N.
For instance, if sulfur is present in the primary chain, we consider replacing it with SO and SO\textsubscript{2}.
To refine the selection, we perform substructure searches in PubChem \cite{PubChem22} for candidate `functional group family' (\ie, chemically similar groups). Families with fewer than approximately 1000 occurrences in PubChem were not included. In addition, we added a set of halogen substituents and organometallic groups relevant to extreme ultraviolet (EUV) photoresists to the functional group list.
Examples of functional groups are shown in \autoref{fig:substructures}, and a visualization of their diversity is illustrated in Supplementary Figure 1. 
Our list is among the most extensive available in the open-source domain \cite{Ertl2017,Ertl2020,Salmina2015-sj,Manelfi2024}, yet its scope remains limited to widely occurring substructures and to organic chemistry.

For carbon backbones, we include standard combinations of three to six carbon atoms with single, double, or triple bonds, as well as common three- to six-membered rings.
If a carbon backbone substructure is fully contained within another substructure, we retain only the larger one to minimize redundancy.
We defined functional groups and carbon backbones substructures to ensure that they are sufficiently overlapping. As discussed in the following section, this overlap will allow to create a more distinctive fingerprint.
Additionally, in Supplementary Note 1, we support the relevance of our substructures by evaluating their coverage on 122M molecules from PubChem. 

\subsection{Substructure-Graph Construction}

In this section, we present how the detected substructures are used to construct a substructure-graph. 
The nodes of the substructure-graph represent the identified substructures and its edges correspond to the intersections between these substructures.

Substructures are considered intersected if their predicted bounding boxes overlap. To enhance connectivity, each detected box is expanded by a margin. This expansion ensures that adjacent substructures sharing a single atom are linked. Additionally, substructures that are not directly connected in the molecular graph may still form connections in the substructure-graph if they are spatially close in the depicted molecular conformation. This added connectivity incorporates positional information, improving the representativity of the substructure graph. 
In the implementation, the expansion of each bounding box is set to 10\% of the diagonal length of the smallest detected box in the image. This expansion influences which groups are overlapping. Supplementary Table 1 shows that the similarity between resulting fingerprints is minimally sensitive to this parameter. A relatively small expansion is preferred to limit the overall size of the fingerprint.

\subsection{Substructure-based Visual Molecular Fingerprint}

To describe molecules and Markush structures, SubGrapher constructs a visual molecular fingerprint by combining detected substructures. This fingerprint encodes the occurrences of functional groups and carbon segments, as well as their distances.

We introduce the Substructure-based Visual Molecular Fingerprint (SVMF). For a given molecule $m$, SVMF$(m)$ is defined as an upper triangular matrix:

\vspace{-5mm}
\begin{equation}
\text{SVMF}(m) = 
\begin{bmatrix}
& \vspace{-2mm} \ddots & & & & & & & \\ 
& & & \hspace{2mm} f_{i, i} & \hspace{2mm} \dots & \hspace{2mm} g_{i, j} & & & \\ 
& & & & \vspace{-2mm} \ddots & \hspace{2mm} \vdots & & & \\ 
& & & {\text{\fontsize{250}{300}\selectfont 0}} & & \hspace{2mm} f_{j, j} &  & & \\ 
& & & & & & \ddots& & \\
 \end{bmatrix}
\end{equation}

$\text{SVMF(m)} \in \mathbb{R}^{n \times n}$, with $n$ is the total number of substructures considered.  Here, $n$ is equal to 1561, including both functional groups and carbon backbone substructures. The diagonal elements $f_{i, i}$ represent the substructure coefficients and are defined as:

\vspace{-5mm}
\begin{equation}
f_{i, i} = h_1 . n_{i} + g_{i, i}
\end{equation}

where $n_{i}$ is the number of instances of substructure $i$ in the molecule, $h_1$ is a model hyperparameter and $g_{i, i}$ is defined as below. The off-diagonal elements $g_{i, j}$ represent the intersection coefficient between substructures $i$ and $j$:

\vspace{-5mm}
\begin{equation}
g_{i, j} = \sum_{\alpha=1}^{n_i} \sum_{\beta=1}^{n_j} h_2\circ d(i_{\alpha}, j_{\beta})
\end{equation}
\vspace{-5mm}

where $\circ$ is the composition operator, $d(i_{\alpha}, j_{\beta})$ denotes the distance between instances $i_{\alpha}$ and $j_{\beta}$, and $n_i$ and $n_j$ are the number of instances of the substructures $i$ and $j$.
The distance $d(i_{\alpha}, j_{\beta})$ is defined as the smallest number of substructures required to construct a path between instances $i_{\alpha}$ and $j_{\beta}$. 
The function $h_2$ is an hyper-parameter of the model controlling the number of non-zero intersection coefficients, and ultimately the fingerprint compression. The SVMF is count-based, as each coefficient depends on the number of substructure occurrences, and continuous, since the values are real numbers.

In the implementation, $h_1$ is set to 10, while $h_2(d)$ takes values of $2$, $2$, $2/4$, $2/16$, and $2/256$ for distances $d = 0, 1, 2, 3, 4$, respectively. If $d > 4$, the intersection coefficient is set to zero. For carbon chain instances, intersection coefficients are divided by $2$ to assign greater significance to functional groups. This prioritization reflects the intuition that functional groups are the most important parts of the molecule.
The SVMF, which has a dimension $1561\times1561 = 1\times2436721$, is not stored in its dense form. Instead, only the non-zero coefficients and their corresponding values are encoded in a compressed vector. The average number of non-zero coefficients for the evaluated fingerprints is reported in the Supplementary Table 2. For the SVMF, the average proportion of non-zero coefficients is $0.001$\%.

\subsection{Molecule Retrieval}

The similarity between two fingerprints of molecules $m_1$ and $m_2$, $f(m_1)$ and $f(m_2)$, can be computed as:

\begin{equation}
    s(f(m_1), f(m_2)) = 
    \frac{\parallel f(m_1) - f(m_2)\parallel}{\parallel f(m_1) + f(m_2)\parallel}
\end{equation}

where $\parallel .\parallel$ denotes the Euclidean norm.
This measure enables searching in large datasets by comparing a query molecule or Markush structure with others to retrieve structurally similar structures.
Further details are in the Results section below.

\subsection{Training Dataset Generation}

Datasets of molecule and Markush structure images extracted from real documents are scarce. Especially, there is no dataset which includes mask annotations. To address this limitation, we use the synthetic data generation pipeline introduced in MolDepictor \cite{Morin_2023_ICCV}, which generates a diverse set of molecular depictions. This pipeline selects molecular SMILES from PubChem and renders them using the RDKit library \cite{RDKit}.

As this pipeline was originally designed to generate molecular graphs for supervision, we extend it to produce mask annotations for detected substructures. To generate an image for each substructure within a molecule, we first identify substructures using RDKit. Next, we post-process the generated SVG images, mapping SVG elements to the corresponding atoms and bonds within each substructure.

Additionally, we improve the pipeline to generate depictions of Markush structures, including structural, positional, and frequency variation indicators. For this purpose, we first construct artificial Markush structures represented as CXSMILES \cite{cxsmiles}, modifying SMILES from PubChem by replacing atom labels with R-group labels, attaching functional groups or R-groups to rings, and including bracket notations. Finally, we use the CDK library \cite{Willighagen2017} to generate Markush structure depictions along with their corresponding substructure mask annotations.

\section{Results and Discussion}

In this section, we perform experiments to evaluate SubGrapher for substructure detection and fingerprinting. Our method is evaluated against state-of-the-art OCSR models and SMILES-based fingerprinting methods.

\subsection{Substructure Detection Evaluation}

\subsubsection{Datasets and Metrics}

To compare our method with state-of-the-art approaches for substructure detection, we evaluate it on three benchmarks: JPO \cite{ChemInfty}, a subset of USPTO-10K-L \cite{Morin_2023_ICCV}, and USPTO-Markush \cite{Morin_2025_CVPR}. These datasets contain pairs of images and structures stored as MOL files \cite{MolFile}.  

JPO consists of 341 molecule images extracted from patent documents published by the Japanese Patent Office \cite{JPO}. This dataset is challenging due to non-standard drawing conventions and low image quality. Molecules containing abbreviations were removed from the standard JPO benchmark set as SubGrapher is currently unable to recognize functional groups in abbreviations. USPTO-10K-L contains 10000 images of large molecules with over 70 atoms and no abbreviations. We evaluate our method using the first 1000 images of this benchmark. USPTO-Markush includes 74 images of Markush structures from patent documents published by the United States Patent and Trademark Office (USPTO \cite{USPTO}).  

MOL files \cite{MolFile} contain the atoms, bonds and their connectivity, allowing us to extract the ground-truth substructures present in a molecule. The substructures evaluated are the 1534 functional groups and the carbon backbones are not considered. 

To evaluate performances, we compute the Substructure F1-score (S-F1) and the Molecule Exact Match (M-EM). The substructure F1-score measures recall (the proportion of ground-truth substructures correctly identified by predictions) and precision (the proportion of predicted substructures that correctly match the ground truth) for substructure detection. The molecule exact match is the percentage of molecules where all substructures are perfectly detected, \ie, the S-F1 equals one.

SubGrapher is compared with three OCSR methods used for substructure detection: MolGrapher, OSRA, and DECIMER. These methods have the advantage of being more general, as converting images to SMILES enables the detection of any functional group, including those outside the scope of our evaluation set. However, OSRA and DECIMER do not predict atoms positions, therefore our evaluation metrics consider only the presence or absence of substructures, not their exact location. 

\subsubsection{State-of-the-art Comparison}

\begin{table*}
\vspace{15mm}
\centering
\resizebox{0.9\linewidth}{!}{
\begin{tabular}{lccccccccc}
\toprule
\multirow{2}{*}{\textbf{Methods}} &  & \multicolumn{2}{c}{JPO (341)} &  & \multicolumn{2}{c}{USPTO-10K-L (1000)} &  & \multicolumn{2}{c}{USPTO-Markush (74)} \\ \cmidrule(lr){3-4} \cmidrule(lr){6-7} \cmidrule(lr){9-10} 
 &  & S-F1 & M-EM &  & \hspace{6mm}S-F1 & M-EM &  & \hspace{6mm}S-F1 & M-EM \\ \midrule
\textit{Rule-based methods} \\
OSRA \cite{OSRA} &  & 81 & 67 &  & \hspace{7mm}\textbf{97} & \textbf{75} &  & \hspace{6mm}74 & 70 \\
\midrule
\textit{Learning-based methods} \\
DECIMER \cite{Rajan2023} &  & 86 & 79 &  & \hspace{7mm}86 & \underline{66} &  & \hspace{6mm}10 & 11 \\
MolScribe \cite{Qian2023} &  & \textbf{94} & \underline{82} &  & \hspace{7mm}90 & 55 &  & \hspace{6mm}\underline{86} & \textbf{86} \\
MolGrapher \cite{Morin_2023_ICCV} &  & 89 & 80 &  & \hspace{7mm}56 & 31 &  & \hspace{6mm}35 & 30 \\
\textbf{SubGrapher (Ours)} &  & \underline{92} & \textbf{83} &  & \hspace{7mm}\textbf{97} & 55 &  & \hspace{6mm}\textbf{88} & \underline{82} \\ \bottomrule
\end{tabular}}\vspace{5mm}
\caption{\textbf{Substructures detection comparison.} Evaluation on benchmarks of molecule images (JPO, USPTO-10K-L) and Markush structure images (USPTO-Markush). Substructure F1-score (S-F1) evaluates the recall and precision for substructure detection. Molecule Exact Match (M-EM) evaluates the percentage of molecules where all substructures are perfectly detected, \ie, the S-F1 equals one. Best scores are bold and second-best scores are underlined.}
\label{tab:substructure-eval}
\end{table*}

\begin{figure*}
    \centering
    \vspace{-7mm}
    \includegraphics[trim={2.5cm 5.5cm 2.5cm 5.5cm}, clip, width=0.95\textwidth]{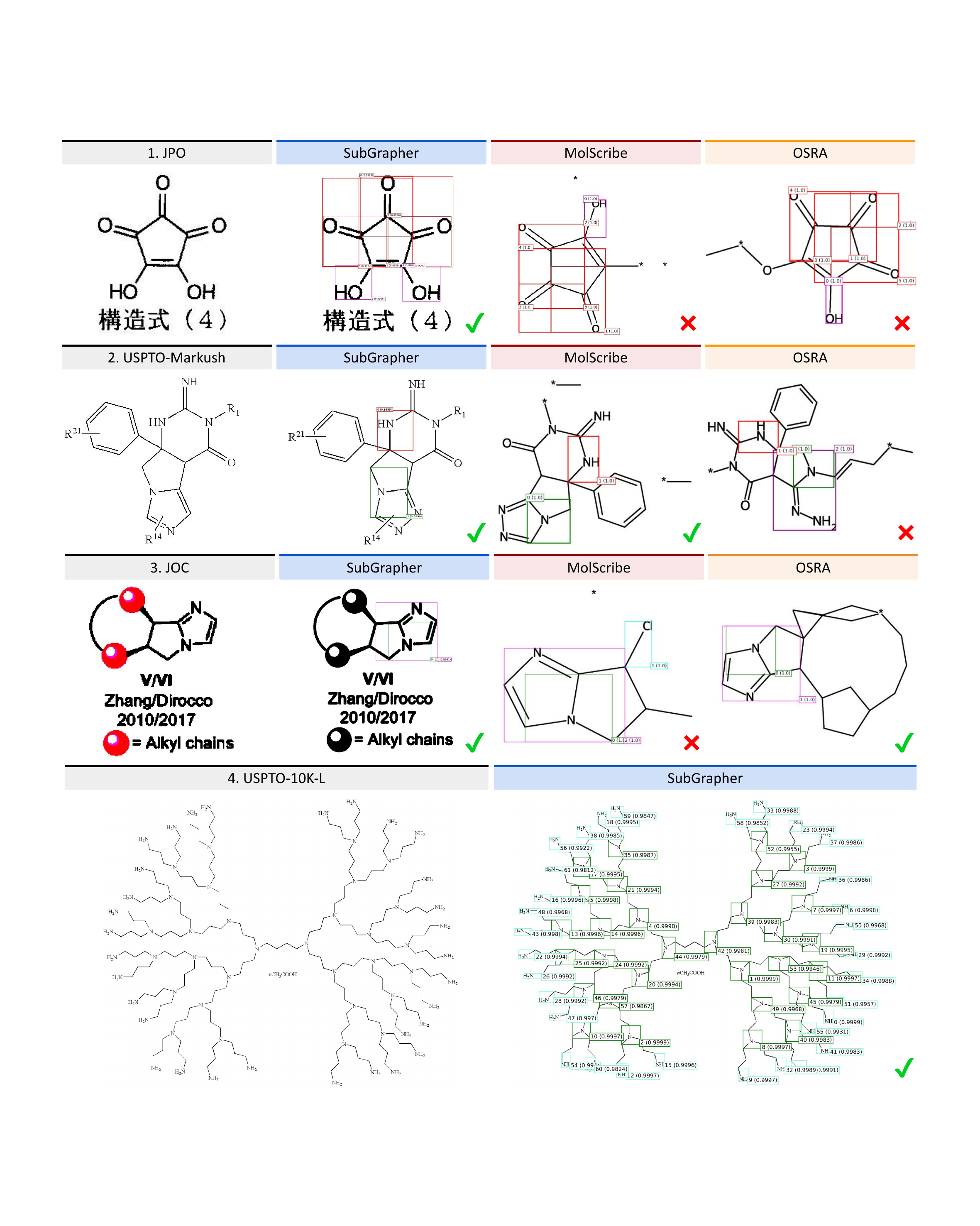}\vspace{3mm}
    \caption{\textbf{Substructure detection qualitative evaluation.} Examples of predicted functional groups are shown for images from patent documents (JPO, USPTO-10K-L, USPTO-Markush) and a scientific journal (JOC).}
    \label{fig:qualitative_eval}\vspace{-3mm}
\end{figure*}

\autoref{tab:substructure-eval} compares SubGrapher with state-of-the-art methods for substructure detection. 
Our method achieves the highest molecule exact match on the JPO dataset. It demonstrates its robustness to the lower-quality images commonly found in this dataset.

On the USPTO-10-L dataset, SubGrapher outperforms other deep-learning approaches in substructure F1-score. Unlike models such as MolGrapher and MolScribe, which experience a performance drop with larger molecules, SubGrapher maintains consistent performance. This suggests that, compared to OCSR models, object detectors are better equipped to handle variations in image scale. Our method is also supervised using a stronger supervision than standard OCSR models, specifically pixel-level (mask) annotations rather than supervision from SMILES strings or molecular graphs. This enables improved recognition of fine details in images of large molecules.

On the USPTO-Markush dataset, SubGrapher again outperforms other methods in substructure F1-score. 
This superior performance results from the limitations of competing models: OSRA and MolGrapher are trained exclusively on molecular images, whereas MolScribe and DECIMER handle Markush structures but support only limited Markush features, specifically variable groups represented as abbreviations. SubGrapher effectively manages complex Markush structures by focusing on relevant regions of the image and excluding Markush-related annotations.

\subsection{Qualitative Evaluation}

\autoref{fig:qualitative_eval} showcases examples of predicted molecules for images from various benchmark datasets. SubGrapher accurately recognizes functional groups in molecule images that contain captions or are of lower quality (\autoref{fig:qualitative_eval}, row 1). It also recognizes functional groups in complex Markush structures (\autoref{fig:qualitative_eval}, row 2) or unconventional drawings displayed in scientific publications (\autoref{fig:qualitative_eval}, row 3). Additionally, our method maintain strong performances on extremely large molecules (\autoref{fig:qualitative_eval}, row 4). Unlike image captioning methods such as DECIMER, our predictions preserve the spatial arrangement of substructures from the input image. This conveys useful information for human interpretation. Additional examples of predictions and typical failure cases are shown in the Supplementary Figure 2.

\subsection{Visual Fingerprinting Evaluation}

In this section, we evaluate SubGrapher for the retrieval of molecules or Markush structures in a collection of images. 

\begin{figure*}
    \centering
    \includegraphics[trim={3.5cm 2.5cm 3.5cm 3cm}, clip, width=0.9\textwidth]{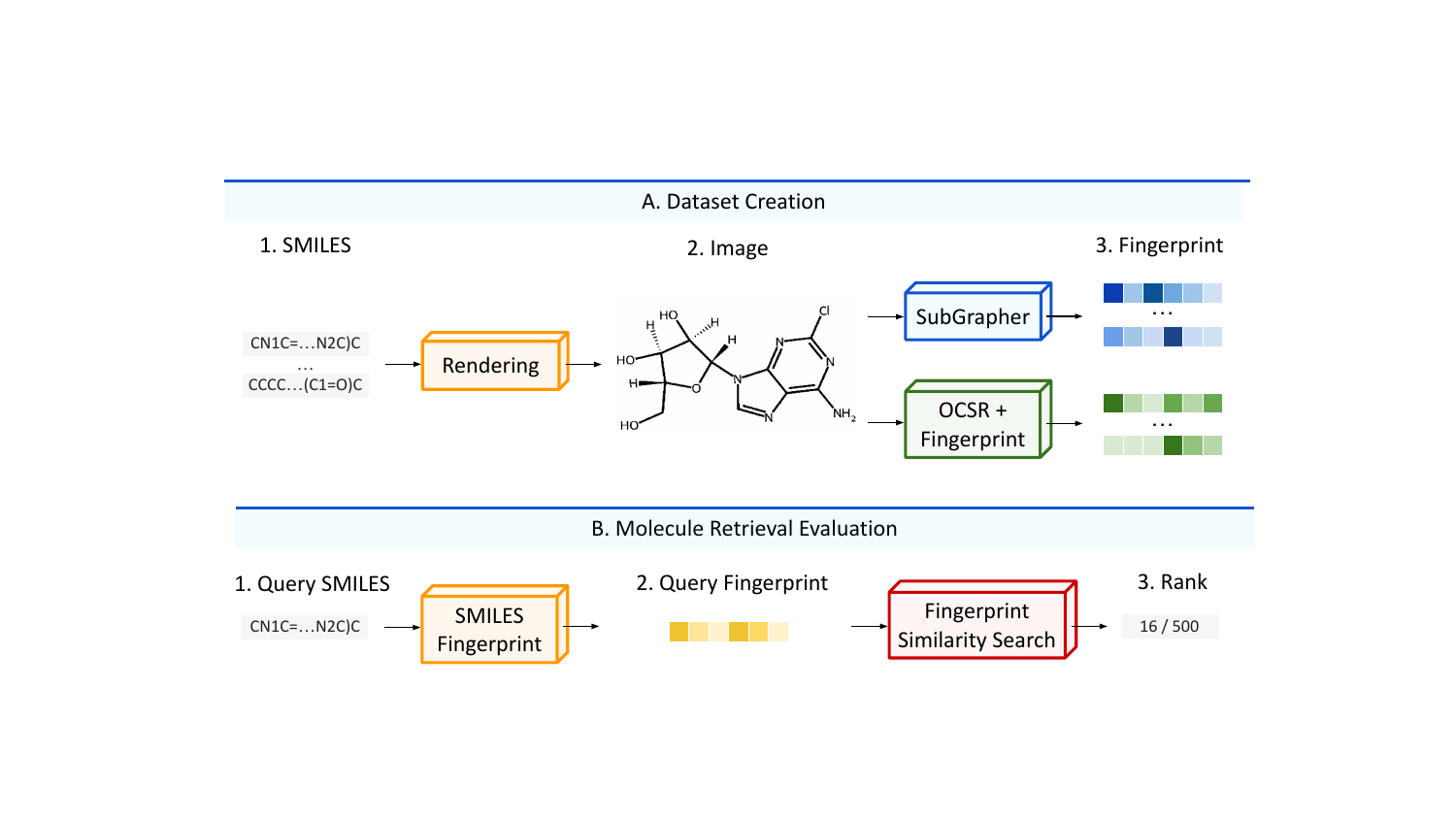}\vspace{5mm}
    \caption{\textbf{Visual fingerprinting evaluation strategy.} (A) First, a set of similar molecules are rendered into images. These images are subsequently converted into fingerprints using each of the evaluated methods. (B) Second, a molecule is converted into a fingerprint based on its SMILES. Its similarity is calculated against all fingerprints within the dataset. Finally, the correct molecule’s position is determined from the ranking of similarity scores.}
    \label{fig:molecule_retrieval}\vspace{-3mm}
\end{figure*}

\subsubsection{Datasets and Metrics}

We evaluate visual fingerprinting methods using five benchmarks of molecules: adenosine, camphor, cholesterol, limonene, and pyridine. The construction of these benchmarks is illustrated in \autoref{fig:molecule_retrieval}.

First, for each reference molecule (adenosine, camphor, cholesterol, limonene, and pyridine), we sample 500 SMILES from PubChem that have at least 90\% structural similarity to the corresponding reference molecule. Here, the structural similarity is the Tanimoto \cite{tanimoto1958elementary} similarity computed between the PubChem fingerprints.
Second, we convert this sets of similar molecules to a sets of images by rendering them using RDKit. The resulting images undergo substantial data augmentation, including scaling, rotation, downscaling, and grid distortion. 
Third, each visual fingerprinting method converts these augmented images into molecular fingerprints. The methods evaluated are SubGrapher, OSRA with RDKit Daylight \cite{RDKit} or MHFP \cite{Probst2018}, and MolScribe with RDKit Daylight or MHFP.
Then, to assess retrieval performance, we measure each method's ability to correctly retrieve a molecular image from a provided SMILES query.
First, the query SMILES is converted into a fingerprint by following the same method than the evaluated fingerprint. Second, the similarity between this query fingerprint and all fingerprints in the dataset is computed. Third, the results are ranked based on similarity scores, and the position of the correct ground-truth molecule is recorded. 
RDKit fingerprint similarity is computed using the Tanimoto similarity on binary fingerprints, the MHFP similarity is computed using the Jaccard similarity, and the SVMF similarity is computed using the Euclidean distance.
Each benchmark is tested using 50 SMILES queries, and the average rank across these queries is reported. 
This setup aims to evaluate the ability to retrieve depictions of a molecule in a large collection of documents based on a provided SMILES query.
As all molecular images in the dataset are highly similar and augmented, despite the relatively small size of test sets, we aim to simulate searches in large collections. Example images from the benchmarks are shown in the Supplementary Figure 3.

\subsubsection{State-of-the-art Comparison}

\begin{table*}
\centering
\resizebox{0.85\linewidth}{!}{
\begin{tabular}{llccccc}
\toprule
\multicolumn{2}{l}{\textbf{Models}} & \multirow{2}{*}{\begin{tabular}[c]{@{}c@{}}Adenosine\\ (500)\end{tabular}} & \multirow{2}{*}{\begin{tabular}[c]{@{}c@{}}Camphor\\ (500)\end{tabular}} & \multirow{2}{*}{\begin{tabular}[c]{@{}c@{}}Cholesterol\\ (500)\end{tabular}} & \multirow{2}{*}{\begin{tabular}[c]{@{}c@{}}Limonene\\ (500)\end{tabular}} & \multirow{2}{*}{\begin{tabular}[c]{@{}c@{}}Pyridine\\ (500)\end{tabular}} \\
OCSR & Fingerprint &  &  &  &  &  \\ \midrule
OSRA \cite{OSRA} & RDKit \cite{RDKit} & 184 & 148 & 165 & 219 & 210 \\
MolScribe \cite{MolScribe} & RDKit & 217 & 363 & \underline{149} & 272 & 203 \\ \midrule
OSRA & MHFP \cite{Probst2018} & 139 & \underline{104} & 181 & 152 & \underline{114} \\
MolScribe & MHFP & \textbf{101} & 287 & 187 & \textbf{48} & 283 \\
\multicolumn{2}{c}{\textbf{SubGrapher (Ours)}} & \underline{110} & \textbf{73} & \textbf{106} & \underline{81} & \textbf{103} \\ \bottomrule
\end{tabular}}\vspace{5mm}
\caption{\textbf{Visual fingerprinting comparison.} We compare the retrieval performance of various Optical Chemical Structure Recognition (OCSR) and fingerprinting methods on image datasets generated from adenosine, camphor, cholesterol, limonene, and pyridine. We report the average rank at which a query molecule is retrieved across 50 queries. Best scores are bold and second-best scores are underlined.} 
\label{tab:search-eval}
\end{table*}

\autoref{tab:search-eval} compares visual fingerprinting methods for molecular retrieval. SubGrapher ranks first for datasets derived from camphor, cholesterol, and pyridine, and second for adenosine and limonene.
In average over all benchmarks, SubGrapher retrieves the correct reference molecule at rank 95, significantly outperforming other methods. Given the challenging nature of the evaluation, we expect that, for a search in a large document collection, the target molecule image should appear at worse within the first 100 results. This reasonable number makes manual inspection feasible for critical search applications such as freedom-to-operate or prior-art search \cite{Ohms2021}.
One reason for SubGrapher’s superior performance lies in how it handles uncertain predictions. For OCSR-based methods, all images converted to invalid SMILES are mapped to the same fingerprint. These identical and uninformative fingerprints degrade the ranking quality. In contrast, SubGrapher’s one-stage approach generates a distinctive fingerprint for each prediction, even when the prediction is uncertain.
Overall, SubGrapher performs more consistently across different reference molecules than other methods, though its performance is slightly weaker on adenosine, cholesterol, and pyridine. For adenosine and cholesterol, this may be due to their larger size compared to the other compounds. For pyridine, the lower performance may come from the presence of charged molecules with multiple fragments in the dataset. SubGrapher currently does not recognize any substructures in single-atom fragments, leading to a loss of information.
Besides, in the Supplementary Figure 4, we show that SubGrapher can distinguish positional isomers and homologous compounds, but not enantiomers. SubGrapher currently has the limitation of ignoring stereo-chemistry information. The Supplementary Table 3 also evaluate the impact of the augmentation level of the benchmarks on the SubGrapher performances. 
Overall, our results show that converting molecular images directly into fingerprints in a single step produces more distinctive fingerprints.

\subsubsection{Qualitative Evaluation}

\begin{figure*}
    \centering
    \vspace{-7mm}
    \includegraphics[trim={0cm 17cm 0cm 17cm}, clip, width=\textwidth]{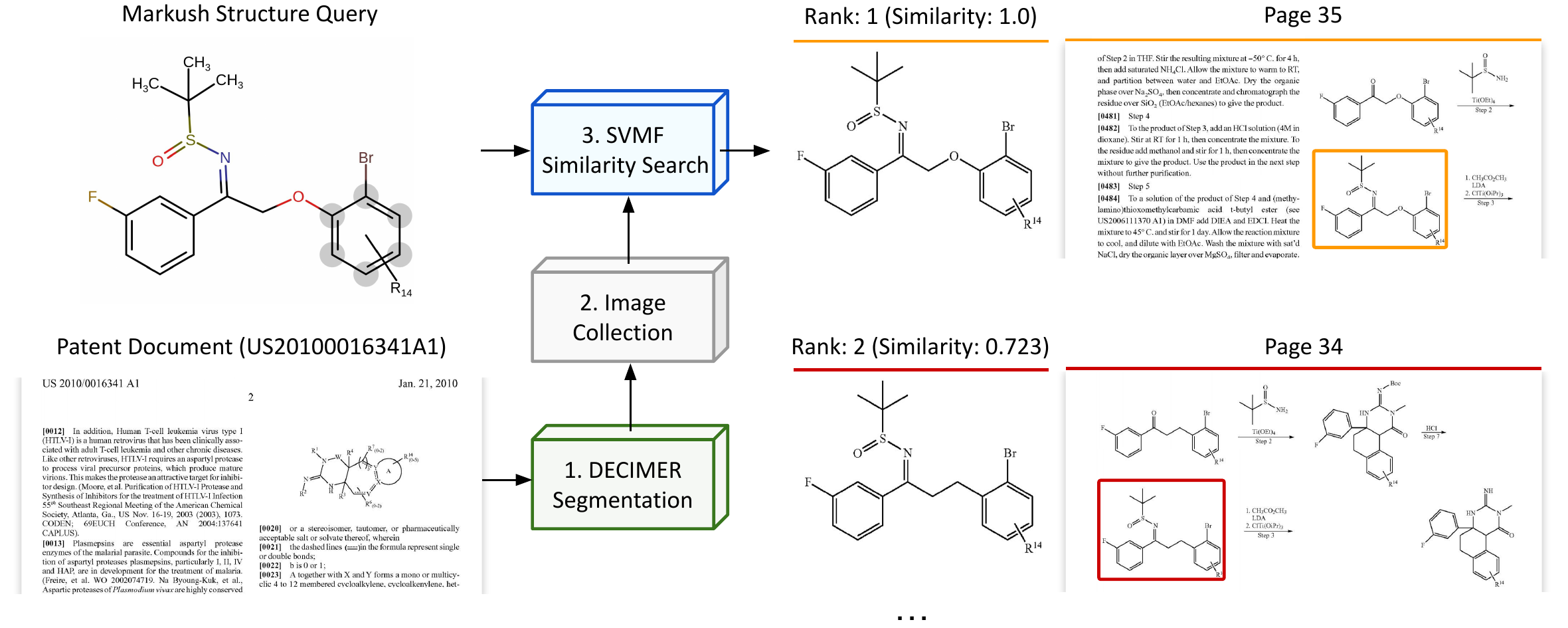}\vspace{3mm}
    \caption{\textbf{Visual fingerprinting qualitative evaluation.} First, images are extracted from a patent document using a molecule image segmentation model. Then, these images are converted to SVMF using SubGrapher. Next, we select a query Markush structure and obtain the fingerprint associated to its structure. Finally, this query fingerprint is compared to all visual fingerprints of the dataset to obtain a ranking.}
    \label{fig:markush_eval}\vspace{-3mm}
\end{figure*}

Here, we demonstrate a use case of SubGrapher for retrieving a Markush structure within a patent document, as illustrated in \autoref{fig:markush_eval}.

First, we extract molecule and Markush structure images from the patent `US20100016341A1' using the DECIMER-Segmentation model \cite{Rajan2021}, a deep neural network designed to segment chemical images from page images. Applying this model to the 54 pages of the document yields 356 extracted images.
Next, each image is converted to its SVMF using SubGrapher. We then randomly select one of the extracted Markush structures and obtain its CXSMILES \cite{cxsmiles} using MarvinJS \cite{Marvin}. The CXSMILES is converted into its associated SVMF using ground-truth matched substructures.

Finally, we compare the query fingerprint against all SVMF in the dataset and rank the results by similarity. \autoref{fig:markush_eval} presents the top two matches along with their corresponding pages from the document. The ground-truth Markush structure image is correctly retrieved as the first match.
This type of search can be particularly valuable for searching information in patent documents that goes beyond standard molecule depictions.

\section{Conclusion}

We present SubGrapher, a learning-based approach for converting 2D depictions of molecules and Markush structures into fingerprints. SubGrapher detects functional groups and carbon backbones, assembles them into a graph representation, and then converts this graph into a fingerprint. This fingerprint enables efficient substructure search and retrieval from collections of molecule and Markush structure images. 
Although trained solely on synthetic images, SubGrapher demonstrates strong generalization to real-world data. Our results highlight the advantages of pixel-level mask supervision for chemical image recognition. Unlike most existing methods, which first convert images to SMILES strings and then derive fingerprints from those SMILES, SubGrapher performs this conversion in a single step. 
While SMILES extraction remains essential for many applications, we show that a single-step approach achieves superior performance for retrieving molecules from image collections. SubGrapher also represents a step towards the retrieval of Markush structures from documents.

\section*{Declarations}

\subsection{Availability of data and materials}

\noindent The SubGrapher code is available on GitHub: \url{https://github.com/DS4SD/SubGrapher/}. 

\noindent The SubGrapher model weights are available on HuggingFace: \url{https://huggingface.co/ds4sd/SubGrapher}.

\noindent The visual fingerprinting benchmarks are available on HuggingFace: \url{https://huggingface.co/datasets/ds4sd/SubGrapher-Datasets}.

\subsection{Authors' contributions}

L.M., G.I.M. and V.W conceptualized the substructure recognition model and fingerprinting method. L.M implemented the code. P.W.J.S and L.V.G supervised the work. L.M. wrote a draft of the manuscript. All authors revised and commented on the manuscript.

\newcommand*{\dictchar}[1]{%
  \clearpage
  \onecolumn
  \begin{center}#1\end{center}
}

\dictchar{{\fontsize{13.5}{16}\selectfont \textbf{Supplementary Information:\\
SubGrapher: Visual Fingerprinting of Chemical Structures}}}

\begin{figure*}[b]
    \centering
    \includegraphics[trim={0cm 0cm 0cm 0cm}, clip, width=\textwidth]{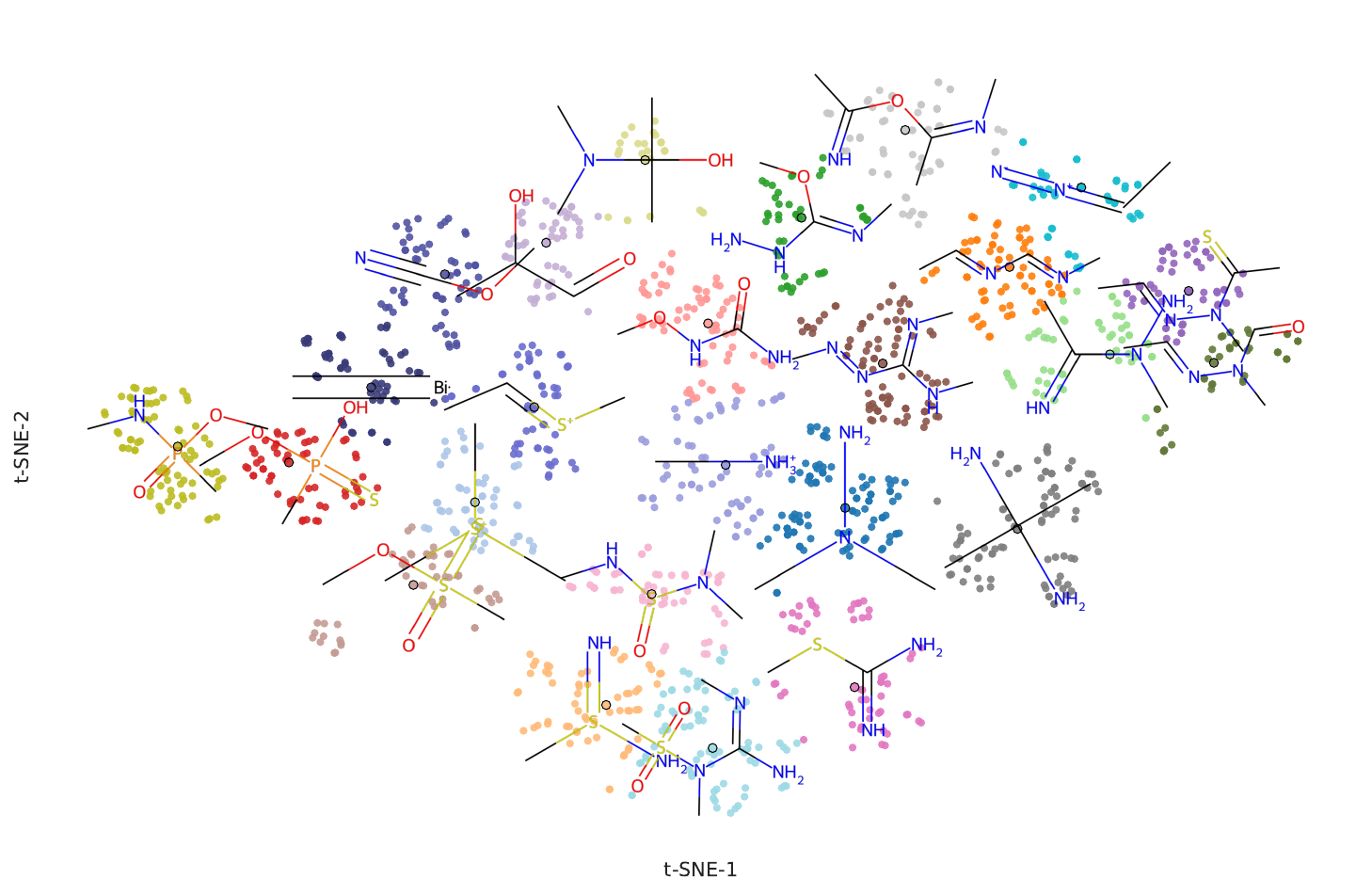}\vspace{5mm}
    \caption{\textbf{Functional groups diversity.} t-SNE visualization of 25 functional group clusters. Each cluster is represented by colored points and overlaid with the centroid molecule.}
    \label{fig:tsne_clusters}
\end{figure*}

\subsection{Supplementary Note 1}

To assess the coverage of our substructures, we analyzed their presence in 122M molecules from PubChem. First, we evaluated how well our functional groups cover the heteroatoms in molecules that contain at least one heteroatom. We found that 97\% of such molecules include at least one of our functional groups. On average, 77\% of their heteroatoms are covered by at least one detected functional group.
Second, we performed the same analysis for carbon backbones in molecules composed solely of carbon atoms. We found that 95\% of these carbon-only molecules include at least one of our carbon patterns. On average, 71\% of their atoms are captured by at least one carbon backbone from our list.

\begin{table}[t]
\centering
\resizebox{0.85\linewidth}{!}{
\begin{tabular}{lccccc}
\toprule
Expansion value & Adenosine & Camphor & Cholesterol & Limonene & Pyridine \\ \midrule
$-1*d$ & 125 & 80 & 126 & 89 & 112 \\
$0*d$ & 111 & 74 & 106 & 82 & 105 \\
$0.05*d$ & 111 & 74 & 106 & 81 & 103 \\
$0.1*d$ & 110 & 73 & 106 & 81 & 103 \\
$0.25*d$ & 110 & 73 & 106 & 82 & 104 \\ \bottomrule
\end{tabular}}\vspace{5mm}
\caption{\textbf{Expansion value sensitivity analysis.} We compare the retrieval performance of SubGrapher for various expansion values of the detected bounding boxes on image datasets generated from adenosine, camphor, cholesterol, limonene, and pyridine. We report the average rank at which a query molecule is retrieved across 50 queries. $d$ denotes the diagonal length of the smallest detected box in the image.}
\label{tab:expansion}
\end{table}

\begin{table}[]
\centering
\resizebox{\linewidth}{!}{
\begin{tabular}{lccccccc}
\toprule
\multirow{2}{*}{Fingerprint} & \multirow{2}{*}{Type} & \multirow{2}{*}{Dimension} & \multirow{2}{*}{Adenosine} & \multirow{2}{*}{Camphor} & \multirow{2}{*}{Cholesterol} & \multirow{2}{*}{Limonene} & \multirow{2}{*}{Pyridine} \\
 &  &  &  &  &  &  &  \\ \midrule
RDKit \cite{RDKit} & Binary & $1\times4096$ & 1465 & 392 & 696 & 230 & 249 \\
HMFP \cite{Probst2018} & Integer & $1\times2048$ & 2048 & 2048 & 2048 & 2048 & 2048 \\
\textbf{SVFP (Ours)} & Float & \begin{tabular}[c]{@{}c@{}}$1561\times1561 = $ \\ $1\times2436721$ \end{tabular} & 43 & 20 & 35 & 15 & 7 \\ \bottomrule
\end{tabular}}\vspace{5mm}
\caption{\textbf{Fingerprints characteristics comparison.} Comparison of fingerprints types, dimensions, and average number of non-zero coefficients computed on image retrieval benchmarks generated from adenosine, camphor, cholesterol, limonene, and pyridine. For RDKit and HMFP, the fingerprints are obtained using OSRA predictions.}
\label{tab:sparsity}
\end{table}

\begin{figure*}
    \centering
    \includegraphics[trim={1cm 5cm 1cm 5cm}, clip, width=\textwidth]{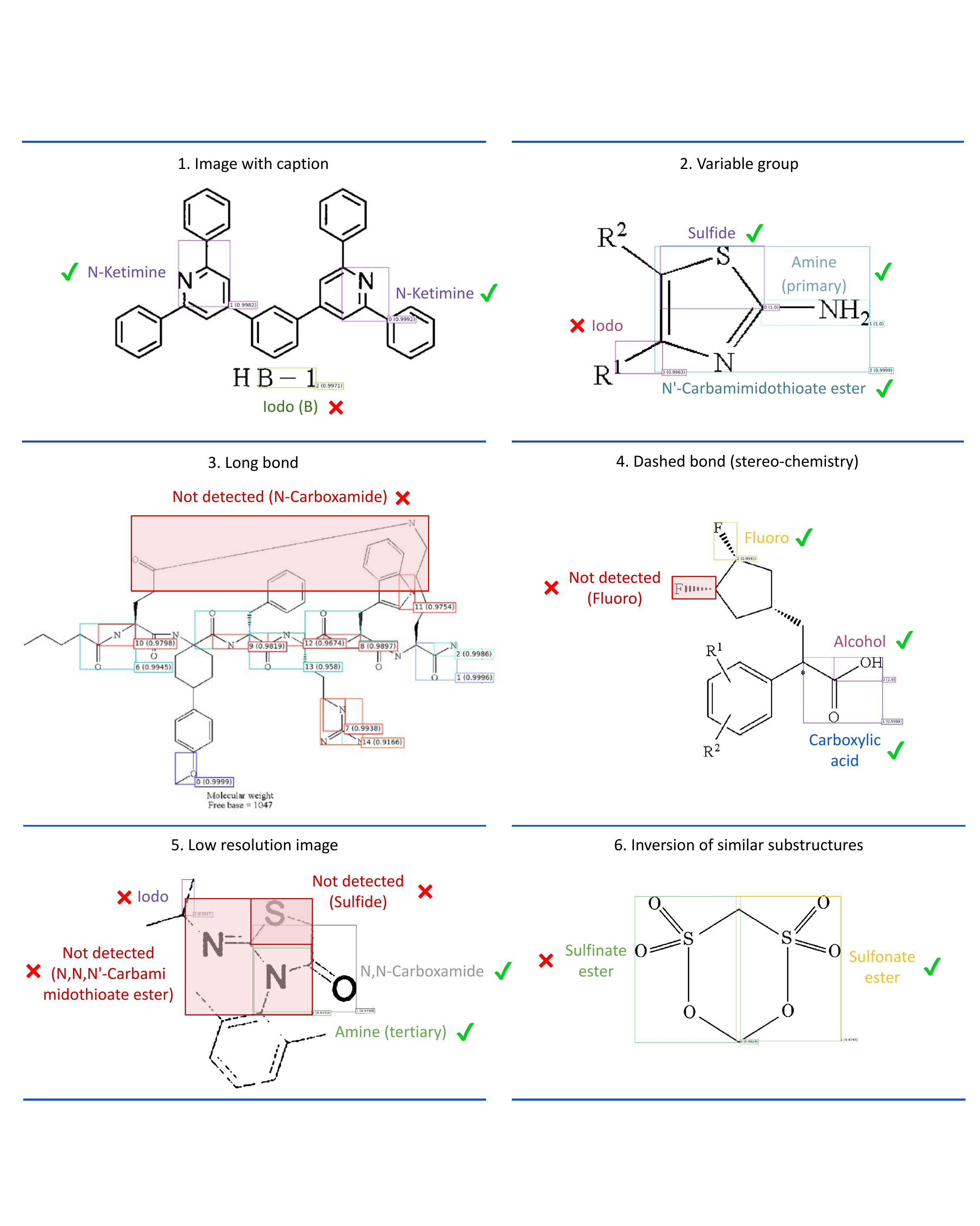}\vspace{2mm}
    \caption{\textbf{Failure cases.} Example of failure cases of SubGrapher on real-world data from JPO, USPTO-Markush and USPTO-10K-L. Typical failure cases include images containing captions (input 1), variable groups (input 2), long bonds (input 3), dashed bonds indicating stereo information (input 4), low resolution images (input 5) and inversions of similar substructures (input 6).}
    \label{fig:fails}
\end{figure*}

\begin{figure*}
    \centering
    \includegraphics[trim={3cm 0cm 3cm 0cm}, clip, width=0.90\textwidth]{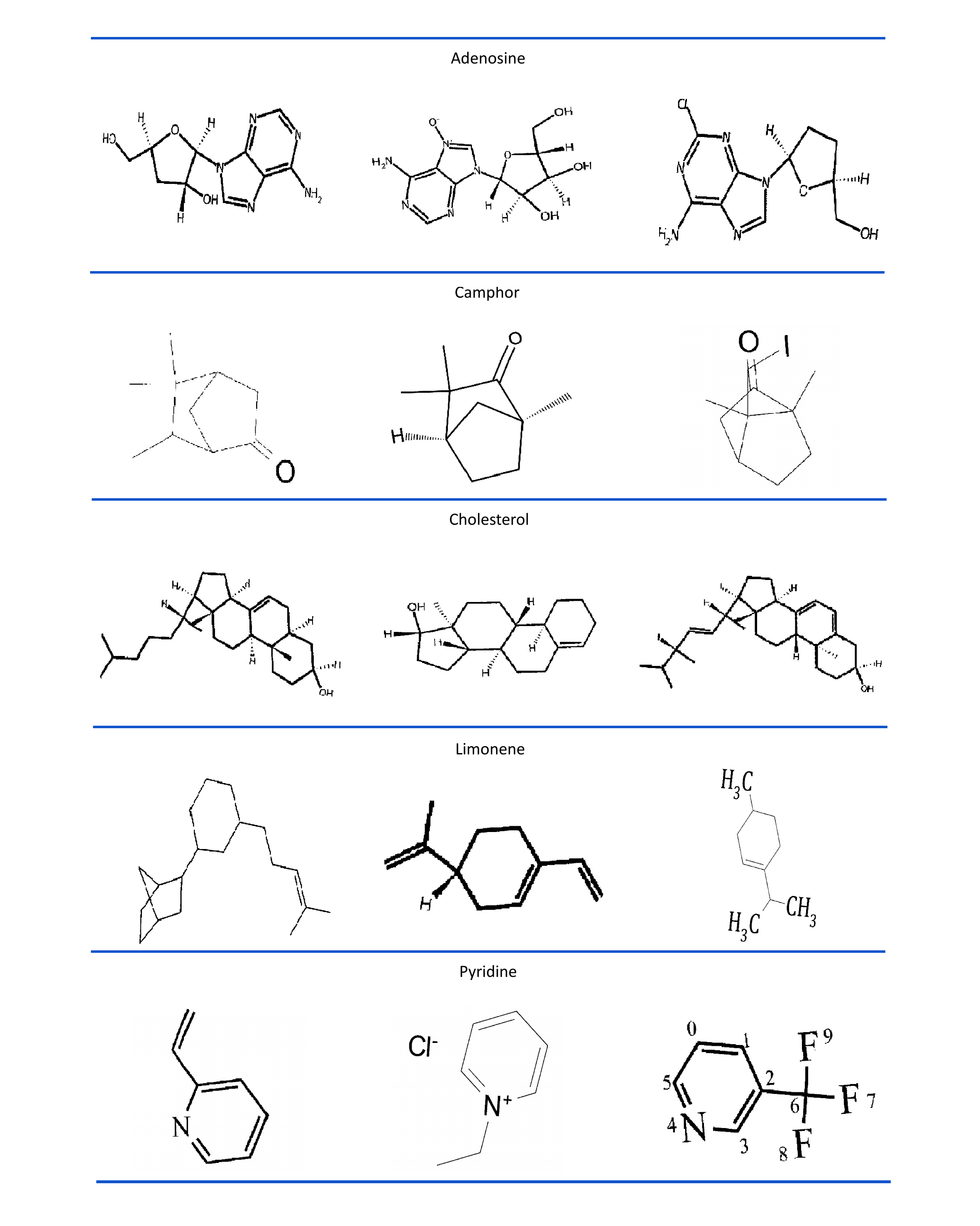}\vspace{-5mm}
    \caption{\textbf{Benchmarks example images.} Example images randomly selected from the benchmark sets generated from adenosine, camphor, cholesterol, limonene, and pyridine.}
    \label{fig:benchmarks}
\end{figure*}

\begin{figure*}
    \centering
    \includegraphics[trim={2cm 9cm 2cm 9cm}, clip, width=\textwidth]{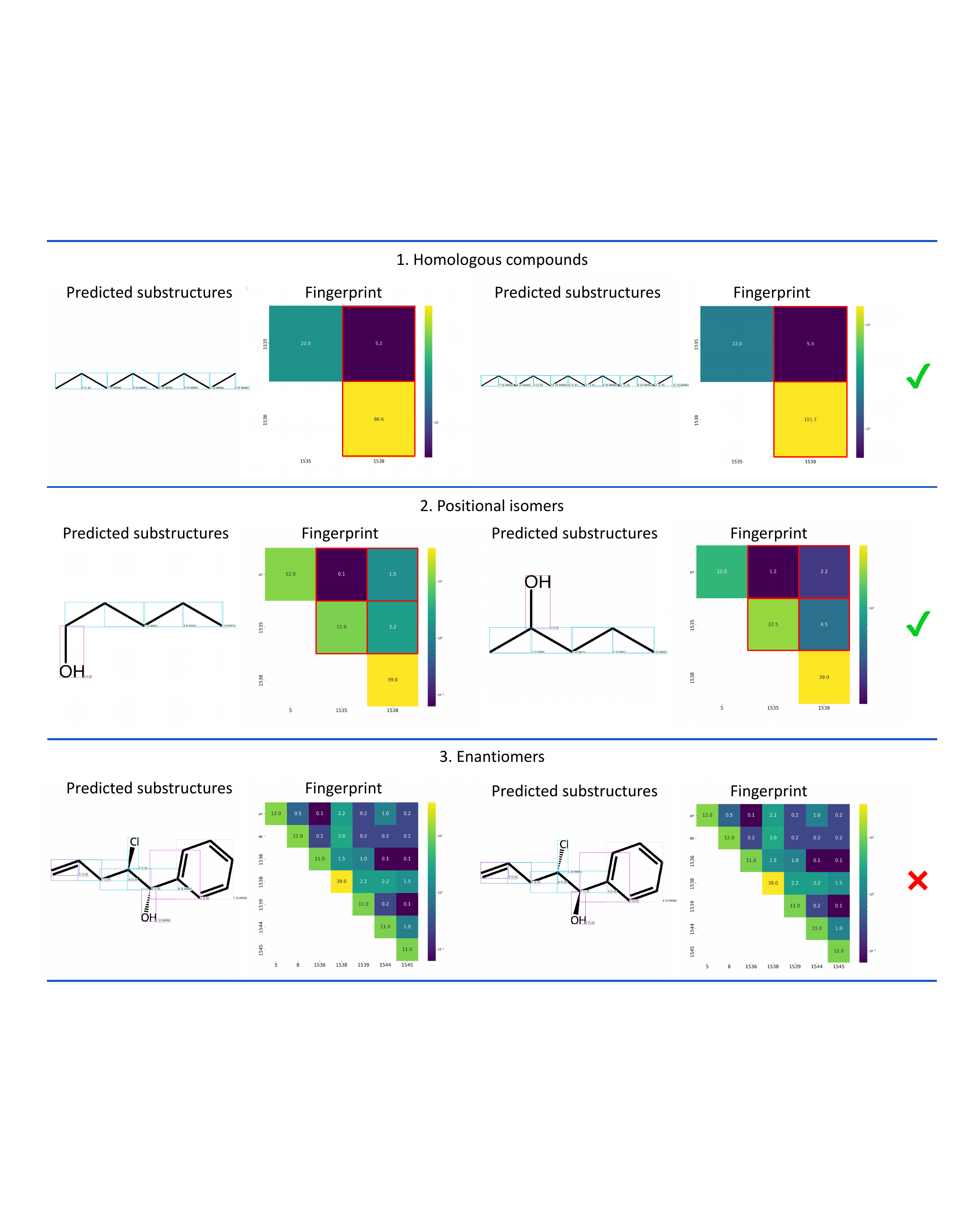}\vspace{2mm}
    \caption{\textbf{SVMF fingerprint discriminative capacity.} SubGrapher's predictions and SVMF fingerprints for example images representing (1) homologous compounds, (2) positional isomers, and (3) enantiomers. Cells in the SVMF fingerprints that differ between compounds are highlighted in red. SubGrapher distinguishes homologous compounds and positional isomers, but not enantiomers, since the substructures it detects lack stereochemistry information.}
    \label{fig:qhomo_enan_pos}
\end{figure*}

\clearpage

\begin{table}[]
\centering
\resizebox{0.9\linewidth}{!}{
\begin{tabular}{lccccc}
\toprule
Augmentation level & Adenosine & Camphor & Cholesterol & Limonene & Pyridine \\ \midrule
Level 1 & 52 & 50 & 76 & 60 & 86 \\
Level 2 & 110 & 73 & 106 & 81 & 103 \\
Level 3 & 159 & 86 & 157 & 85 & 171 \\ \bottomrule
\end{tabular}}\vspace{2mm}
\caption{\textbf{Benchmarks augmentation analysis.} We compare the retrieval performance of SubGrapher on different variants of the image datasets generated from adenosine, camphor, cholesterol, limonene, and pyridine. Each variant uses different augmentations levels as described in the Supplementary Note 2. We report the average rank at which a query molecule is retrieved across 50 queries.}
\label{tab:augmentation_analysis}
\end{table}

\subsection{Supplementary Note 2}

To analyze the molecule retrieval results presented in the main manuscript, we perform evaluations on multiple augmented versions of the benchmarks. 
\autoref{tab:augmentation_analysis} compares the retrieval performance
of SubGrapher on different variants of the image datasets generated from adenosine, camphor, cholesterol, limonene, and pyridine. 
Each benchmark is augmented using different levels of augmentations. 
Level 1 corresponds to applying the augmentations:
\begin{itemize}
    \renewcommand\labelitemi{-}
    \item Rotation with factor drawn between -0.1 and 0.1 (applied with a probability of 90\%),
    \item Scaling with factor drawn between -0.4 and -0.3 (90\%), 
    \item Downscaling with a factor drawn between 0.5 and 0.8 (70\%),
    \item Grid distortion with factor drawn between -0.1 and 0.1 (50\%).
\end{itemize}
Level 2 is used in the main manuscript and corresponds to applying the augmentations:
\begin{itemize}
    \renewcommand\labelitemi{-}
    \item Rotation with factor drawn between -0.1 and 0.1 (90\%),
    \item Scaling with factor drawn between -0.7 and -0.5 (90\%), 
    \item Downscaling with a factor drawn between 0.7 and 0.99 (70\%),
    \item Grid distortion with factor drawn between -0.15 and 0.15 (50\%).
\end{itemize}
Level 3 corresponds to applying the augmentations:
\begin{itemize}
    \renewcommand\labelitemi{-}
    \item Rotation with factor drawn between -0.15 and 0.15 (90\%),
    \item Scaling with factor drawn between -0.8 and -0.6 (90\%), 
    \item Downscaling with a factor drawn between 0.8 and 0.99 (70\%),
    \item Grid distortion with factor drawn between -0.2 and 0.2 (50\%).
\end{itemize}

We observe that augmentation, particularly strong downscaling and grid distortion, has a significant impact on performance. This effect is especially pronounced for the benchmarks derived from adenosine, cholesterol, and pyridine. An explanation is that these benchmarks contain more heteroatoms than the limonene and camphor benchmarks, making them more sensitive to the loss of detail caused by downscaling.

\bibliography{main}

\end{document}